\documentclass{article}
\usepackage[table]{xcolor}
\usepackage{graphicx}
\graphicspath{ {./resources/} }
\usepackage{pdfpages}
\usepackage{caption}
\usepackage{subcaption}
\usepackage{float}
\usepackage{amsmath}
\usepackage{authblk}
\usepackage{hyperref}
\hypersetup{
  colorlinks,
  citecolor=blue,
  linkcolor=blue,
  urlcolor=blue,
}

\title{Hawk: An Industrial-strength Multi-label Document Classifier}
\author{Arshad Javeed}
\affil{
Lund University, Sweden

email: ar1886ja-s@student.lu.se
}
\date{}

\begin{document}
\maketitle

\section{Abstract}

There are a plethora of methods and algorithms that solve the classical multi-label document classification. However, when it comes to deployment and usage in an industry setting, most, if not all the contemporary approaches fail to address some of the vital aspects or requirements of an ideal solution: i. ability to operate on variable-length texts and rambling documents. ii. catastrophic forgetting problem. iii. modularity when it comes to online learning and updating the model. iv. ability to spotlight relevant text while producing the prediction, i.e. visualizing the predictions. v. ability to operate on imbalanced or skewed datasets. vi. scalability. The paper describes the significance of these problems in detail and proposes a unique neural network architecture that addresses the above problems. The proposed architecture views documents as a sequence of sentences and leverages sentence-level embeddings for input representation. A hydranet-like architecture is designed to have granular control over and improve the modularity, coupled with a weighted loss driving task-specific heads. In particular, two specific mechanisms are compared: Bi-LSTM and Transformer-based. The architecture is benchmarked on some of the popular benchmarking datasets such as Web of Science - 5763, Web of Science - 11967, BBC Sports, and BBC News datasets. The experimental results reveal that the proposed model outperforms the existing methods by a substantial margin. The ablation study includes comparisons of the impact of the attention mechanism and the application of weighted loss functions to train the task-specific heads in the hydranet.

\section{Introduction \& Related Work}

Document classification has been a historical and classical machine-learning problem. Multi-label document classification involves assigning a subset of possible labels to the given document. Today, there are numerous methods that solve the task offering unique features, ranging from primitive methods such as SVMs, ensemble methods, and boosting methods, to convoluted state-of-the-art methods, deep neural networks - CNNs, LSTMs, and transformer-based networks.\cite{survey_text_clf, survey_comparative_study} have compared a broad spectrum of contemporary methods and benchmarked the performance. Document classification has gained traction as it is also employed as a preprocessing step prior to other downstream tasks such as clustering, topic modeling, and dialog extraction for Q\&A. The more broader application domain of text classification in the industry is data governance: a technology-assisted review of documents in the context of legality, sensitivity, personally-identifiable information, and other relevant labels or classes relevant to a specific organization \cite{tar_exp, tar_review}. But when it comes to the deployment or usage of a machine-learning model in a real-world application, there are more aspects than just a set of metrics that ultimately define the robustness and usability of the solution. The paper describes some of the vital desirable characteristics of an ``ideal document classifier'' and then alludes to the design countering the problems.

One of the key requirements is the ability to operate on variable-length and rambling documents. Often, document classifiers of varying architectures are trained to classify texts of an ideal length that is chosen by looking at the training dataset, and the assumption is that production data is similar in nature. The fundamental flaw here is that the machine-learning models are not scale-invariant. A CNN can detect patterns of similar size or scale it was trained \cite{multi_scale_order}. Mechanisms to circumvent the problem include, increasing the receptive field by widening the network \cite{imagenet_clf_deep, very_deep_cn}. Or learning features at multiple scales \cite{scale_inv_conv_nn}. But such mechanisms are not feasible when dealing with text data, where data augmentation is not possible. \cite{overfit_trans} describe the performance of a transformer on inconsistent sequences. Another drawback in the case of transformer-based models is the computational overhead. Longformer \cite{longformer}, ToBERT \cite{tobert}, and CoLTX \cite{cogltx} variants of BERT to support long sequences, but computational cost is still high when dealing with large documents instead of sentences or paragraphs and may result in too many features for the fine-tuning layer to deal with. \cite{eff_clf_long_docs} have analyzed the performance of transformer-based models on long documents.

Online updating of models after deployment is quite common these days, as more and more data becomes available. This brings us to the ``catastrophic forgetting problem''. Typically, a multi-label classifier has several hidden layers and an output layer with as many nodes as the number of classes. In a practical scenario, the data may be available in any form: documents that are partially or pseudo-labeled examples, or examples with missing labels. In such scenarios, it is not quite straightforward to update the model with partially labeled data. Simply freezing a few of the nodes in the output layer and training the model can lead to instability and the layers forgetting the things learned. One of the approaches is to do a pseudo-rehearsal \cite{pseudo_rehearsal}, where some amount of training data is retrained and appended to the new data during the update. \cite{overcome_cat} suggest slowing down a few parts of the network during the update process. Although both approaches are viable, they cannot be exploited frequently, and do not guarantee stability. Also, having a single bottleneck hinders the modularity of the network. The liberty to add or remove classes or labels without impacting the performance of the rest of the labels is not possible. So the proposed architecture employs a hydranet architecture \cite{hydranet} allowing specific heads to be trained while completely freezing other branches, guaranteeing stability. Each head in the hydranet may be responsible for a specific task, in our case each head performs a binary classification or a multi-class classification (if a particular set of labels are correlated). The modular architecture also facilitates scalability in terms of parallel or distributed training. As the heads in the hydranet are independent of one another, albeit forming a single model, the distributed training would be significantly faster as gradient updates are simplified.

When classifying a rambling document, it may be desirable to have the model highlight specific portions of text that the model thinks are relevant, instead of just having a set of scores (for the respective classes) for the entire document. This feature is crucial in applications such as data governance where there are extensive (manual) inspections to filter out sensitive data. Several attempts have been made to solve the problem. The prominent approaches involve a CNN model \cite{text_clf_cnn, cnn_sen} that leverages the spatial information from the word-level embeddings. The activation maps would then reveal the relevant portions of text. However, it can be hard to discern and attend to the correct activation map, as each filter in the CNN produces its own activation and the interpretability diminishes as we move deeper into the network. CNNs for text classification also suffer from other limitations that have been discussed above.

To address the above problems, the proposed method views a document as a sequence of sentences of fixed length and operates on sentence-level embeddings to perform sequence classification. The hydranet architecture provides flexibility in terms of training, better stability, and other advantages as described. Two mechanisms, in particular, are explored for the hydranet heads, Bi-LSTM, and transformer-based. The heads are driven by weighted loss functions to better handle imbalanced classification. The ability of the model to highlight the relevant portions of text is studied by looking at the heat maps of time-truncated predictions. Finally, the results are benchmarked with the popular datasets for multi-class and multi-label classification.

\section{Methodology}

\subsection{Input Representation}

In contrast to most of the contemporary methods, the proposed methodology treats documents as a sequence of sentences or blocks of texts of some fixed length (in terms of characters). The motivation for such a choice is the fact that long documents, unlike sentences or short documents, are free-flowing texts. Treating every single word as a feature and attending to it might be redundant, while it might be just sufficient to simply attend to portions of text with a bird's eye view. This would also significantly reduce the computational cost.

There are several pretrained \& opensource models that can be leveraged for sentence-level transformation and embeddings: \cite{sup_learn_usr} propose a sentence encoder using recurrent network networks with attention heads and hierarchical convolutional networks. The universal sentence encoder \cite{use} has two variants, a deep averaging network (DAN) and a transformer-based architecture. This work, however, uses the DAN variant of the universal sentence encoder (USE) for input representation.

Given long text or document $x$, the text is broken down into a number of blocks based on the hyperparameter $s_b$ that represents the size of a particular block in characters. So the document can now be represented as a sequence (equation \ref{eq:input}). Each $b_i$ is a section of text of length $s_b$, and $n_b$ is another hyperparameter that depends on the maximum allowed length ($L$) per document (number of characters). Thus, $n_b = \frac{L}{s_b}$. And $L$ can be chosen to be arbitrarily large. The input $x$ is then fed to the universal sentence encoder to form the input feature matrix (equation \ref{eq:input_feat}) of size $n_b \times n_d$. With $n_d = 512$ being the embedding dimension of the universal sentence encoder. The output representation is straightforward, with a vector of 1s and 0s representing the targets for classification.

\begin{equation}
	\label{eq:input}
	x = [b_1, \ b_2, \ ..., \ b_{n_b}]
\end{equation}

\begin{equation}
	\label{eq:input_feat}
	X = [B_1, \ B_2, \ ..., \ B_{n_b}]
\end{equation}

However, there is a trade-off associated with the two hyperparameters. The parameter $s_b$ defines the size of the individual block that the USE embeds. A large block size produces a small $n_b$ and results in diluted representation in terms of the embeddings, while a smaller block size produces more accurate embeddings, but results in too many feature vectors which could add a computational overhead.

Another aspect to consider when it comes to batched training is padding. As documents are often of different lengths, an empty string padding (equation \ref{eq:input_pad}) is done and a boolean mask (equation \ref{eq:input_mask}, \ref{eq:input_feat_mask}) is generated for subsequent layers. This is necessary, as USE does not handle masking on its own, and produces a dense vector even for an empty string. So the effective input then becomes $X \odot X_m$ (element-wise multiplication).

\begin{equation}
	\label{eq:input_pad}
	x = [b_1, \ b_2, \ ..., e_{n_b-1}, \ \ e_{n_b}]
\end{equation}

\begin{equation}
	\label{eq:input_mask}
	x_m = \neg \verb!is_empty!([b_1, \ b_2, \ ..., e_{n_b-1}, \ \ e_{n_b}]) = [1, \ 1, \ ..., 0, \ 0]
\end{equation}

\begin{equation}
	\label{eq:input_feat_mask}
	X_m = 
\begin{bmatrix}
x_m \\
x_m \\
... \\
x_m
\end{bmatrix}
\end{equation}

\subsection{Hydranet Architecture}

\begin{figure}[H]
	\centering
	\caption{Hydranet Architecture for Multi-label Classification}
	\begin{subfigure}[b]{0.8\textwidth}
		\caption{Frozen backbone, trainable neck and brances}
		\includegraphics[width=\textwidth]{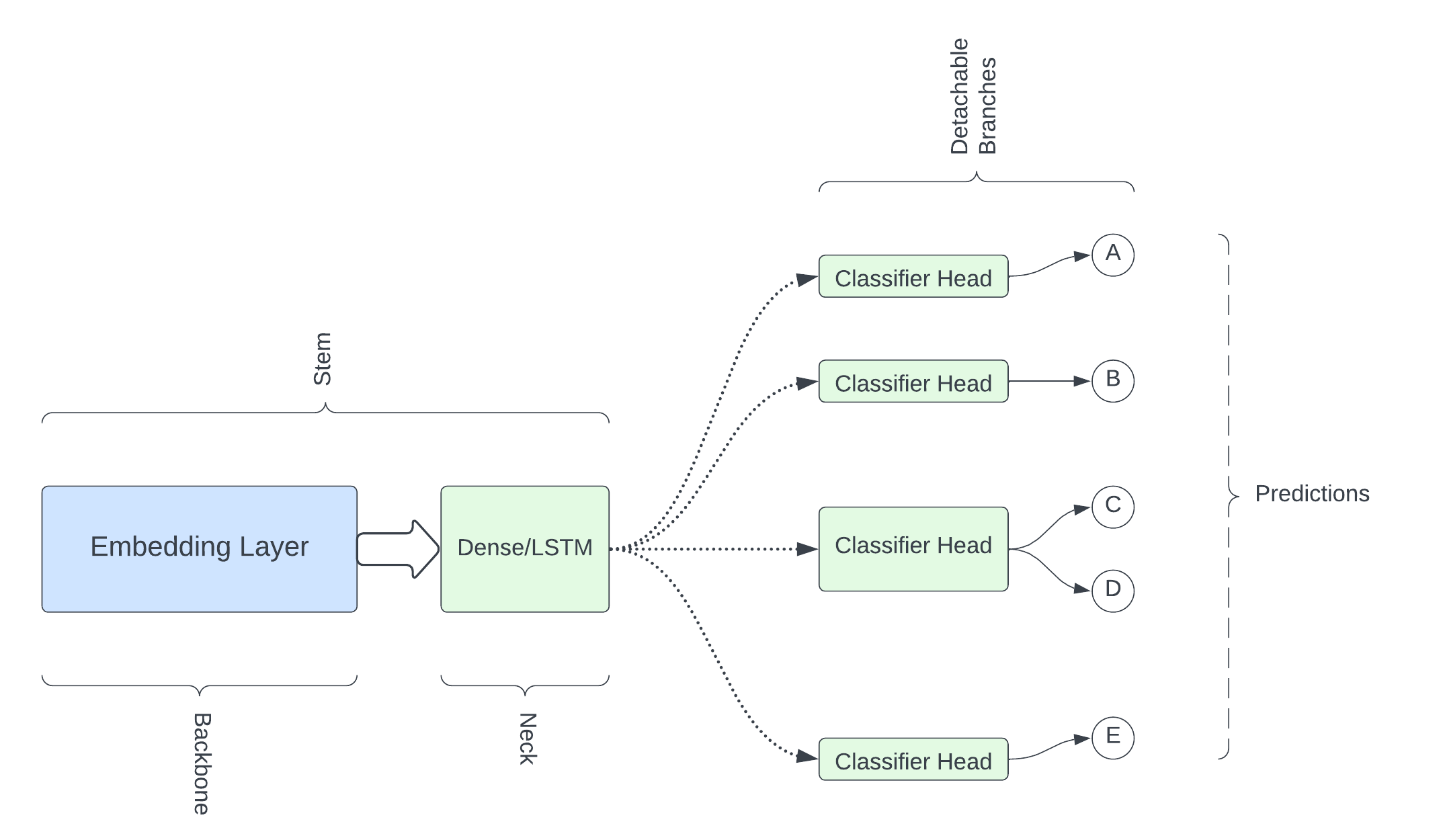}
		\label{fig:hydranet}
	\end{subfigure}
	\hfill
	\begin{subfigure}[b]{0.8\textwidth}
		\caption{Frozen backbone and neck, detached heads}
		\includegraphics[width=\textwidth]{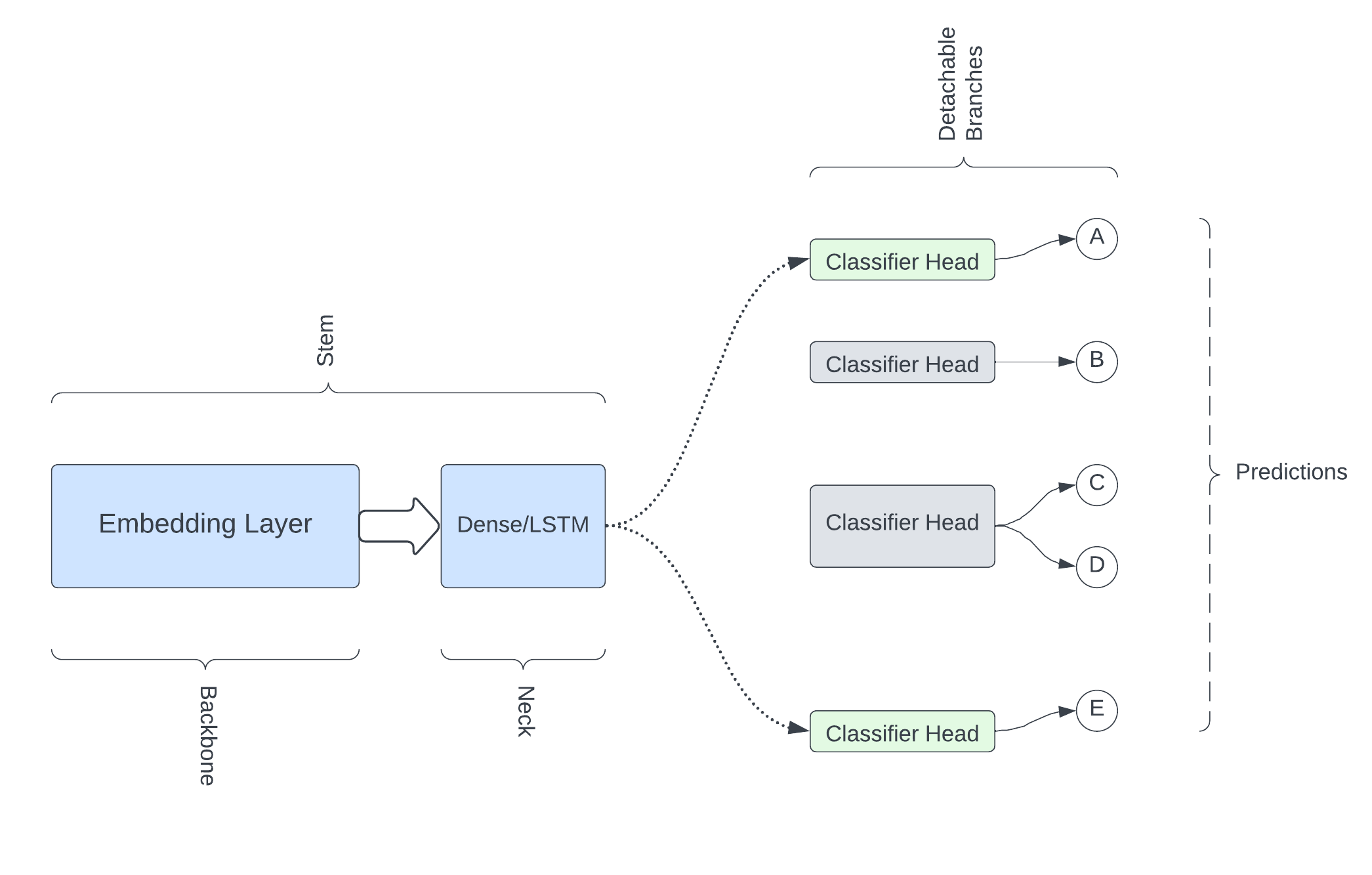}
		\label{fig:hydranet_det}
	\end{subfigure}
\end{figure}

The key to circumventing the catastrophic forgetting problem and gaining the advantages of modularity and scalability is the hydranet architecture. The hydranet architecture \cite{hydranet} consists of a backbone, several task-specific branches, and a gating mechanism that controls the branches to execute. However, the architecture proposed here slightly differs, the gating mechanism is skipped and the branches themselves are allowed to perform the classification. The intention for doing so is to make the branches detachable and modular.  The stem comprises the language embedding model, plus an optional network that can learn common features for all the classes. Figure \ref{fig:hydranet} shows a sample architecture. The heads themselves can be of arbitrary architectures, two specific architectures Bi-LSTM and transformers will be discussed later on. It is worth noting that some heads can perform multi-label classification of a subset of correlated labels so that the information and learnings are shared.

With the defined architecture, there are two ways to perform online learning (after the training phase). If the new data encountered has all the labels marked and can be trusted, then the neck and all of the branches can be trained (similar to the training process). However, if the new data has missing labels, then the respective heads can be detached and the neck can be frozen (figure \ref{fig:hydranet_det}). This will ensure that the performance on other labels stays the same.

\subsection{Bi-LSTM Heads}

The sequence length (in terms of $n_b$) can get large depending on the size of the document. Hence, LSTMs are an ideal choice when it comes to recurrent networks due to the gating their gating mechanisms. And the bi-directionality makes enables better attention.

Equations \ref{eq:lstm_i} - \ref{eq:lstm_h} characterise the LSTM operation. Where the input $X$ is now the masked input, $X \odot X_m$. Since we have a bi-directional LSTM, the sequences from forward and reverse propagation are simply concatenated (equation \ref{eq:lstm_h_fb}). Additionally, the new hidden representations can be fed through a dense layer as a means to compute the significance of each feature or time slice as seen in equation \ref{eq:lstm_a1} - \ref{eq:lstm_a2}, $H$ is the matrix consisting of hidden features, and $\omega$ is the weight matrix of the dense layer. Table \ref{tab:bilstm_head} provides a summary of the Bi-LSTM head architecture.

\begin{equation}
	\label{eq:lstm_i}
	i_t. =\sigma(W_i X + U_i h_{t-1} + b_i)
\end{equation}

\begin{equation}
	\label{eq:lstm_o}
	o_t. =\sigma(W_o X + U_o h_{t-1} + b_o)
\end{equation}

\begin{equation}
	\label{eq:lstm_f}
	f_t. =\sigma(W_f X + U_f h_{t-1} + b_f)
\end{equation}

\begin{equation}
	\label{eq:lstm_cc}
	\hat c_t. =\sigma(W_c X + U_c h_{t-1} + b_c)
\end{equation}

\begin{equation}
	\label{eq:lstm_c}
	c_t = f_t \odot c_{t-1} + i_t \odot \hat c_t
\end{equation}

\begin{equation}
	\label{eq:lstm_h}
	h_t = o_t \odot \sigma(c_t)
\end{equation}

\begin{equation}
	\label{eq:lstm_h_fb}
	\hat h_t = [h_t, \tilde h_t]
\end{equation}

\begin{equation}
	\label{eq:lstm_a1}
	\alpha = \omega^T * \tanh(H)
\end{equation}

\begin{equation}
	\label{eq:lstm_a2}
	\hat H = H * \alpha^T
\end{equation}

\begin{table}[h]
\caption{Bi-LSTM Head Architecture}
	\label{tab:bilstm_head}
	\centering
	\begin{tabular}{|c|c|}
		\hline
		\textbf{Layer} & \textbf{Description} \\
		\hline \hline
		Bi-LSTM &  10 units, ReLU\\
		\hline
		Bi-LSTM &  10 units, ReLU\\
		\hline
		Dense &  10 units, ReLU\\
		\hline
		Dense &  5 units, ReLU\\
		\hline
		Dense &  1 unit, Sigmoid\\
		\hline
	\end{tabular}
\end{table}

\subsection{Transformer Heads}

The transformer heads comprise self-attention heads, pooling layers, and a dense layer for classification. Equations \ref{eq:transformer_a1} - \ref{eq:transformer_a2} summarize the self-attention operation for one particular attention head. $\omega^Q, \omega^K, \omega^O$ are the attention weights. Additionally, there is an attention mask $\alpha_m$  (equation \ref{eq:transformer_a3}) that controls the width of the attention window ($\tau$) and introduces the hyperparameters $\tau$. A matrix of all ones corresponds to global attention, while an eye matrix corresponds to no attention and local attention for intermediate values. Table \ref{tab:transformer_head} provides a summary of the transformer head architecture.

\begin{equation}
	\label{eq:transformer_a1}
	\alpha = \frac{\hat H * \omega^Q \cdot (\hat H * \omega^K)^T}{\sqrt{d_k}} \hat H
\end{equation}

\begin{equation}
	\label{eq:transformer_a2}
	\hat H_{\alpha} = \omega^O * \hat H * (\alpha^T \odot \alpha_m(\tau))
\end{equation}

\begin{equation}
	\label{eq:transformer_a3}
	\alpha_m(\tau = 1) = 
	\begin{bmatrix}
	1 & 1 & 0 & ... & 0 \\
	0 & 1 & 1 & ... & 0 \\
	0 & 0 & 1 & ... & 0 \\
	... & ... & ... & ... & ... \\
	0 & 0 & ... & 1 & 1 \\
	\end{bmatrix}
\end{equation}

\begin{table}[h]
\caption{Transformer Head Architecture}
	\label{tab:transformer_head}
	\centering
	\begin{tabular}{|c|c|}
		\hline
		\textbf{Layer} & \textbf{Description} \\
		\hline \hline
		Self Attention &  3 multi-head attention heads 32 units each, ReLU\\
		\hline
		Pooling &  Average pooling, ReLU\\
		\hline
		Dropout &  $P = 0.1$ \\
		\hline
		Dense &  20 units, ReLU\\
		\hline
		Dropout &  $P = 0.1$ \\
		\hline
		Dense &  1 unit, Sigmoid\\
		\hline
	\end{tabular}
\end{table}

\section{Training}

The input ($X$) to the network is a set of long documents (strings) that are broken down into blocks to form a string matrix of size $n \times n_b$ and the targets ($y$) are one-hot vectors of size $n \times num\_classes$ for individual heads to act upon. As each of the branches acts independently and does the classification, each branch has the liberty to formulate its own loss function. For instance, a regular binary cross-entropy loss, f1-loss, weighted BCE, or categorical cross entropy, in the case of a particular branch is responsible for multiple labels. The proposed work employs a weighted BCE loss to drive individual classification heads.

\begin{equation}
	\label{eq:loss1}
	E = -\frac{1}{N} \Sigma_n \omega_{k0} (1 - y) \log(1 - \hat y) + \omega_{k1} y \log(\hat y)
\end{equation}

\begin{equation}
	\label{eq:loss2}
	\omega_{k0} = \frac{N}{(N - N_k) * K}, \omega_{k1} = \frac{N}{N_k * K}
\end{equation}

Equations \ref{eq:loss1} - \ref{eq:loss2} define the weighted BCE. Where $y, \hat y$ are the true and predicted targets, $\omega_k$ is the weight for class $k$, and $N, N_k$ represent the total number of samples and the number of positive samples for class $k$, and $K$ is the total number of classes. Table \ref{tab:hyperparams} lists the other hyperparameters involved.

\begin{table}[H]
\caption{Training Hyperparameters}
	\label{tab:hyperparams}
	\centering
	\begin{tabular}{|c|c|}
		\hline
		\textbf{Hyperparameter} & \textbf{Value} \\
		\hline \hline
		Embedding dimension (USE) & 512 \\
		\hline
		Max. document length (chars) & 5,000 \\
		\hline
		Block size (chars) & 100 \\
		\hline
		Batch size & 16 \\
		\hline
		Number of epochs & 5 \\
		\hline
	\end{tabular}
\end{table}

\section{Visualizing Predictions}

The idea here is to view the predictions as a heatmap over the document, not just a set of confidences for the entire document. This requires constructing the input in a time-truncated form for a time-distributed prediction. Consider an input document (string), $x$, the input is then formatted for prediction as per equation \ref{eq:input_pred}, where $e$ is an empty string and $n_b$ is the number of blocks. The new input is then fed to the network for a time-distributed prediction of size $n_t \times K$, with $n_t$ being the number of time slices for the input, and $K$ being the number of classes. The obtained prediction matrix reveals the way the network's predictions evolve with the input, thereby making it easy to pin-point the relevant blocks responsible for triggering the network.

\begin{equation}
	\label{eq:input_pred}
	x = 
	\begin{bmatrix}
	b_1 & e_2 & e_3 & ... & e_{n_b} \\
	b_1 & b_2 & e_3 & ... & e_{n_b} \\
	b_1 & b_2 & b_3 & ... & e_{n_b} \\
	... & ... & ... & ... & ... \\
	b_1 & b_2 & b_3 & ... & b_{n_b} \\
	\end{bmatrix}
\end{equation}

To validate the behavior, the proposed model was tested by concatenating two documents/samples of distinct categories. Table 1 lists the sample inputs. Two distinct samples are simply concatenated to create a document belonging to both categories (table \ref{tab:data_viz}, id: 3). It is also important to note that such a ”hybrid” sample was seen by the model during the training phase, and the training dataset consisted of multi-class samples, i.e. each sample belonging to exactly one class. With this setup, the expectation is not only for the model to predict descent probabilities for the two categories but also to identify the texts responsible for the prediction.

\begin{figure}[H]
\centering
\caption{Visualizing Predictions - Bi-LSTM}
\label{fig:lstm_viz}
\begin{subfigure}[b]{0.45\textwidth}
\caption{Final Prediction: Heatmap of final prediction}
\includegraphics[width=\textwidth]{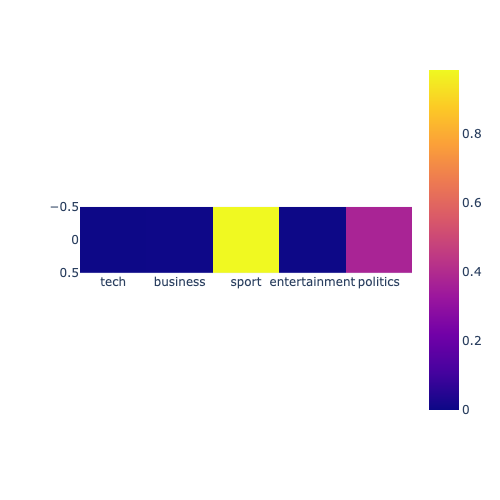}
\label{fig:lstm_viz_p}
\end{subfigure}
\hfill
\begin{subfigure}[b]{0.45\textwidth}
\caption{Time-distributed Prediction (Subsequences): Each row represents a truncated subsequence and the columns represent the labels.}
\includegraphics[width=\textwidth]{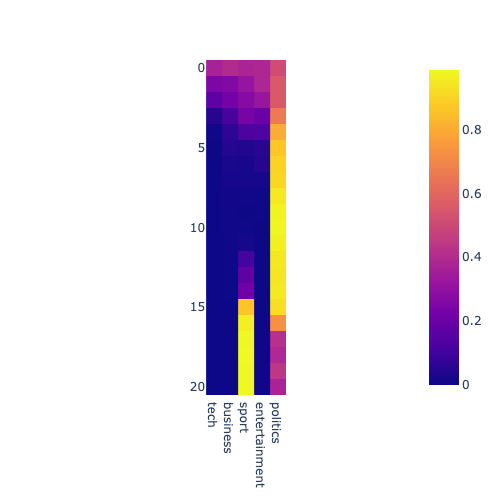}
\label{fig:lstm_viz_pm}
\end{subfigure}
\end{figure}

\begin{figure}[H]
\centering
\caption{Visualizing Predictions - Transformer}
\label{fig:transformer_viz}
\begin{subfigure}[b]{0.45\textwidth}
\caption{Final Prediction: Heatmap of final prediction. Attention window $\tau = 10$}
\includegraphics[width=\textwidth]{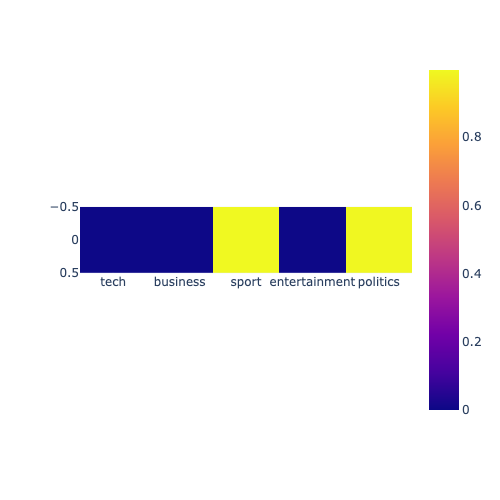}
\label{fig:transformer_viz_p1}
\end{subfigure}
\hfill
\begin{subfigure}[b]{0.45\textwidth}
\caption{Time-distributed Prediction (Subsequences): Each row represents a truncated subsequence and the columns represent the labels. Attention window $\tau = 10$}
\includegraphics[width=\textwidth]{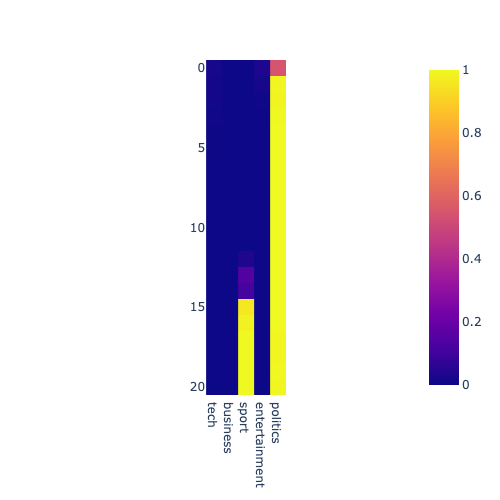}
\label{fig:transformer_viz_pm1}
\end{subfigure}
\hfill
\begin{subfigure}[b]{0.45\textwidth}
\caption{Final Prediction: Heatmap of final prediction. Attention window $\tau = 3$}
\includegraphics[width=\textwidth]{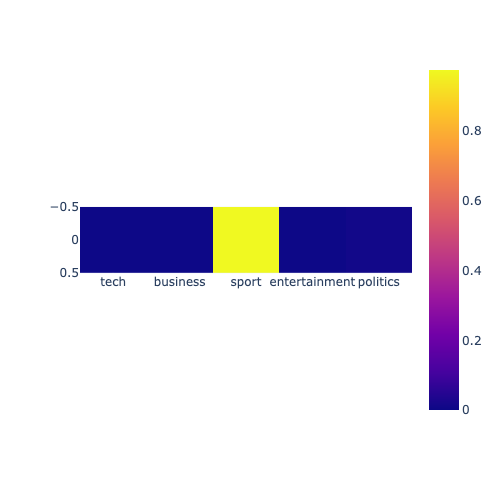}
\label{fig:transformer_viz_p2}
\end{subfigure}
\hfill
\begin{subfigure}[b]{0.45\textwidth}
\caption{Time-distributed Prediction (Subsequences): Each row represents a truncated subsequence and the columns represent the labels. Attention window $\tau = 3$}
\includegraphics[width=\textwidth]{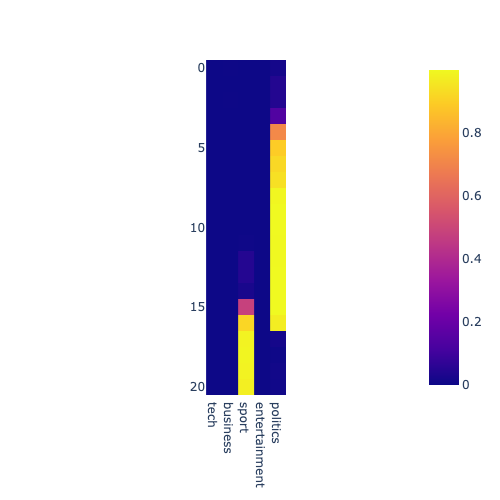}
\label{fig:transformer_viz_pm2}
\end{subfigure}
\end{figure}

Figures \ref{fig:lstm_viz} - \ref{fig:transformer_viz} shows the visualization of the predictions for the LSTM arch and transformer arch respectively. Both the models are able to detect the presence of the two categories in the input (figures \ref{fig:lstm_viz_p} - \ref{fig:transformer_viz_p2}), albeit the transformer predicts the labels with higher probabilities compared to the LSTM model, this is probably due to the global attention mechanism. Inspecting the heatmaps (figures \ref{fig:lstm_viz_pm} - \ref{fig:transformer_viz_pm2}) reveals that the first portion of the document belongs to the category ``politics" and the second half is of ``sports" category. The important but subtle difference in the two maps is that the transformer predicts both categories towards the end due to global attention, while the LSTM's prediction wears off for the other category. It can also be observed that the transformer's predictions have higher confidence compared to the LSTM's.

\begin{table}[H]
\caption{Inputs: Prediction Vizualization}
\label{tab:data_viz}
\centering
\begin{tabular}{|c|p{5cm}|c|c|}
\hline
\textbf{Id} & \textbf{Document} & \textbf{Length (in chars)}& \textbf{Category} \\
\hline
\hline
1 &  blunkett hints at election call ex-home secretary david blunkett ... & 1004 chars/180 words & Politics\\
\hline
2 &  cole faces lengthy injury lay-off aston villa ... & 1006 chars/190 words & Sports\\
\hline
3 & blunkett hints at election ... cole faces lengthy injury lay-off ... & 2010 chars/370 words & Politics, Sports \\
\hline
\end{tabular}
\end{table}

Figures \ref{fig:transformer_viz_p1} - \ref{fig:transformer_viz_pm2} also illustrate the effect of the attention window, $\tau$ (number of blocks). With a larger attention window, $\tau=10$ (figures \ref{fig:transformer_viz_p1} - \ref{fig:transformer_viz_pm1}) the model is able view a larger chunk of the document and is able to retain the prediction for the ``sport'' class even after the relevant text has ended in the first half of the document. In stark contrast, for $\tau = 3$ (figures figures \ref{fig:transformer_viz_p2} - \ref{fig:transformer_viz_pm2}) the predictions for the ``politics'' class wears off similar to Bi-LSTM, albeit the quality of predictions (confidences) are still better. Using the respective attention maps, the relevant texts can be easily extracted. The granularity here again depends on the block size ($s_b$).

\section{Datasets}

Both the variants of the proposed architecture (Bi-LSTM and transfomer heads) are benchmarked on some of the popular datasets for long-text classification: Web of Science - 5763, Web of Science - 11967, BBC Sports, and BBC News. In all cases, the appropriate metrics are adopted for the comparison. In the absence of a predefined test dataset, the metric is reported on 20\% of the random sample (averaged over a few trials). Table \ref{tab:data} summarizes the properties of the datasets. The WOS - 5763 \& WOS - 11967 datasets comprises of published research articles categorized into a predefined set of classes.  WOS - 5763 has 11 labels while the 11967 variant has 33 labels. The BBCS and BBCN dataset contains news articles for topic classification.

\begin{table}[H]
\caption{Datasets}
\label{tab:data}
\centering
\begin{tabular}{|c||c|c|}
\hline
\textbf{Dataset} & \textbf{Number of Labels} & \textbf{Number of Docs} \\
\hline
\hline
Web of Science - 5736 (WOS - 5763) & 11 & 5736 \\
\hline
Web of Science - 11967 (WOS - 11967) & 33 & 11967 \\
\hline
BBC Sports (BBCS) & 5 & 750 \\
\hline
BBC News (BBCN) & 5 & 2225 \\
\hline
\end{tabular}
\end{table}

\section{Results}

For all the experiments the configuration of the hyperparameters is as described in \ref{tab:hyperparams}. Tables \ref{tab:data_wos1} - \ref{tab:data_bbc2} compare the results. We see that across the board, both the Bi-LSTM and the transformer architectures perform quite similarly, albeit Bi-LSTM with a thin edge over the transformer. For the experiments on WOS-5763 and WOS11-967, the proposed architecture outperforms the previous best by a significant margin. On the BBCS dataset, the previous best has an edge of ~ 0.7\%.

\begin{table}[H]
\caption{Comparison on Web of Science - 5763}
\label{tab:data_wos1}
\centering
\begin{tabular}{|c||c|c|}
\hline
\textbf{Model} & \textbf{Accuracy} \\
\hline
DNN \cite{wos_cnn_rnn} & 86.16 \\
CNN \cite{wos_cnn_rnn} & 88.68 \\
RNN \cite{wos_cnn_rnn} & 89.46 \\
XLNet \cite{wos_xlnet} & 90 \\
BERT \cite{wos_bert} & 90.24 \\
HDLTex-CNN \cite{wos_hdl} & 90.93 \\
TM \cite{wos_tm} & 91.28 \\
\hline
\textbf{Proposed (Bi-LSTM)} & \textbf{94.45} \\
\textbf{Proposed (Transformer)} & \textbf{94.00} \\
\hline
\end{tabular}
\end{table}

\begin{table}[H]
\caption{Comparison on Web of Science - 11967}
\label{tab:data_wos2}
\centering
\begin{tabular}{|c||c|c|}
\hline
\textbf{Model} & \textbf{Accuracy} \\
\hline
DNN \cite{wos_cnn_rnn} & 80.02 \\
CNN \cite{wos_cnn_rnn} & 83.29 \\
SVM & 80.65 \\
HDLTex \cite{wos_hdl} & 86.07 \\
\hline
\textbf{Proposed (Bi-LSTM)} & \textbf{92.81} \\
\textbf{Proposed (Transformer)} & \textbf{91.28} \\
\hline
\end{tabular}
\end{table}

\begin{table}[H]
\caption{Comparison on BBC Sports}
\label{tab:data_bbc1}
\centering
\begin{tabular}{|c||c|c|}
\hline
\textbf{Model} & \textbf{Accuracy/micro-F1} \\
\hline
DAN \cite{dan} & 94.78 \\
SPGK \cite {spgk} & 94.97 \\
REL-RWMD \cite{rel_rwmd} & 95.18 \\
ApproxRepSet \cite{aprox_rep} & 95.73 \\
DiSAN \cite{disan} & 96.05 \\
HN-ATT \cite{hntt} & 96.73 \\
WMD \cite{wmd} & 98.71 \\
MPAD-path \cite{mpad} & \textbf{99.59} \\
\hline
\textbf{Proposed (Bi-LSTM)} & 98.88 \\
\textbf{Proposed (Transformer)} & 98.88 \\
\hline
\end{tabular}
\end{table}

\begin{table}[H]
\caption{Comparison on BBC News}
\label{tab:data_bbc2}
\centering
\begin{tabular}{|c||c|c|}
\hline
\textbf{Proposed (Bi-LSTM)} & \textbf{99.06} \\
\textbf{Proposed (Transformer)} & \textbf{98.98} \\
\hline
\end{tabular}
\end{table}

\section{Conclusion}

The objective of the paper was a present a pragmatic solution for document classification suitable for an industrial setting. Vital problems such as document length, catastrophic forgetting problem, modularity, and explainability have been addressed to a good extent with the proposed architecture. The proposed architecture views documents as a sequence of sentences and performs sequence classification using a recurrent architecture (Bi-LSTM) and a transformer architecture. Both methods perform considerably well compared to the baselines. There is further room for experimentation and improvement considering other variants and flavors for the base embedding layer. All the results described in the paper are based on the state-of-the-art Universal Sentence Encoder model (DAN variant), for simplicity and computational efficiency. Other transformer-based models can also be substituted as the base model which may lead to some improvement in the overall accuracy.

\bibliographystyle{apalike}
\bibliography{./resources/bib.bib}

\end{document}